\documentclass[]{spie}

\usepackage{amsmath,amsfonts,amssymb}
\usepackage{graphicx}
\usepackage{multirow}
\usepackage[colorlinks=true, allcolors=blue]{hyperref}

\title{Byte-level generative predictions for forensics multimedia carving}

\author[a]{Jaewon Lee}
\author[b]{Md Eimran Hossain Eimon}
\author[a]{Avinash Srinivasan}
\author[b]{Hari Kalva}

\affil[a]{United States Naval Academy, 121 Blake Road, Annapolis, USA}
\affil[b]{Florida Atlantic University, 777 Glades Road, Boca Raton, USA}

\authorinfo{Further author information: Send correspondence to leejaewon075@gmail.com.\\
    \textbf{Disclaimer:} The views expressed in this document are those of the authors and do not reflect the official policy or position of the U.S. Naval Academy, Department of the Navy, the Department of War, or the U.S. Government.
    }

\begin{document}
\maketitle

\begin{abstract}
Digital forensic investigations often face significant challenges when recovering fragmented multimedia files that lack file system metadata. While traditional file carving relies on signatures and discriminative deep learning models for fragment classification, these methods cannot reconstruct or predict missing data. We propose a generative approach to multimedia carving using bGPT, a byte-level transformer designed for next-byte prediction. By feeding partial BMP image data into the model, we simulate the generation of likely fragment continuations. We evaluate the fidelity of these predictions using different metrics, namely, \textit{cosine similarity, structural similarity index (SSIM), chi-square distance,} and \textit{Jensen-Shannon divergence (JSD)}. Our findings demonstrate that generative models can effectively predict byte-level patterns to support fragment matching in unallocated disk space.
\end{abstract}


\section{INTRODUCTION}\label{sec:intro}
This research addresses the critical challenge of recovering fragmented multimedia data when file system metadata is lost or compromised. To present the problem context and motivate the proposed solution, this introduction is structured in three parts. First, we examine the complexities of modern multimedia forensics and the reliance on file system structures. Second, we discuss the inherent limitations of traditional signature-based and discriminative machine learning carving methods. Finally, we introduce our proposed generative approach using byte-level transformers to predict and reconstruct missing data fragments.

\subsection{The Challenge of Multimedia Forensics}
The increasingly digital world has led to a proliferation of multimedia data stored across many platforms, including smartphones, personal computers, and Internet of Things (IoT) devices\cite{b1}. These multimedia files, such as images, audio, and video, are integral to personal communication, government operations, and commercial applications\cite{b1}. However, files can become partially or fully inaccessible due to hardware failure, human error, intentional criminal tampering, or malicious cyber activity\cite{b2}.

In digital forensic investigations, the recovery of corrupted or deleted files is an essential operation, particularly when multimedia content serves as key evidence in criminal, civil, or national security investigations\cite{b2}. However, when file system metadata, such as \textit{inode tables} in Linux-based systems or the \textit{Master File Table} (MFT) in Windows NTFS, is absent or corrupted, traditional file recovery tools become unreliable\cite{b1}. This challenge is especially significant in unallocated disk space, where deleted or fragmented files exist without file system pointers\cite{b3}.

To overcome this limitation, forensic analysts employ a method known as file carving, which enables data recovery directly from raw disk space by detecting inherent file signatures and structural patterns\cite{b2}. Because file carving does not depend on file allocation metadata, it has become an indispensable tool for digital forensics analysts to reconstruct usable content from corrupted disk space\cite{b2}.

\subsection{Limitations of Traditional Carving Methods}
File carving is conventionally divided into two primary stages: the classification of file fragments and their subsequent reassembly into complete files \cite{b4}. However, traditional carving methods face significant limitations in practical forensic scenarios. In real-world environments, key structural indicators such as file headers and footers may be fragmented or entirely absent \cite{b4}. Furthermore, file fragments are frequently stored non-contiguously on disk, which complicates reassembly efforts and reduces the overall reliability of signature-based detection \cite{b3}.

While recent advances demonstrate the effectiveness of deep learning models, particularly \textit{convolutional neural networks} (CNNs), in accurately identifying file types across large and diverse fragment datasets \cite{b5}, these methods remain fundamentally discriminative. They are designed to classify fragments based on existing patterns but lack the capability to generate or reconstruct missing portions of data when essential segments are lost. This leaves a significant gap in recovery efforts where reassembly is dependent on predicting partial input rather than simply identifying it.

\subsection{The Generative Approach with bGPT}
To address the limitations of discriminative models, this paper introduces a generative approach to multimedia carving by leveraging bGPT, a byte-level transformer model trained for next-byte prediction \cite{b6}. Unlike classification-only methods, bGPT learns from raw binary data to generate plausible byte continuations from a given byte input. Because of this capability, bGPT does not require file signatures or high-level metadata, making it adaptable to scenarios involving severe fragmentation or corruption.

We apply this approach to fragmented BMP images and evaluate the quality of generated fragment predictions against true continuations using similarity metrics such as cosine similarity, chi-square distance, structural similarity index (SSIM), and Jensen-Shannon divergence (JSD). This research explores a new direction in multimedia forensics by combining generative prediction with statistical validation for fragment matching, where missing content can be predicted beyond simple classification. Through this framework, we seek to improve multimedia carving in fragmented and unallocated disk space, where metadata is unavailable but informative byte-level patterns remain.

The remainder of this paper is organized as follows. Section~\ref{sec:rw} provides an overview of related work in multimedia carving and fragment classification, with a particular emphasis on recent machine learning techniques. Section~\ref{sec:preliminaries} establishes the preliminaries of this study, detailing the statistical metrics utilized for similarity comparison and the architecture of the bGPT byte-level transformer. Section~\ref{sec:methodology} describes our four-phase methodology, which includes dataset preparation, generative prediction, similarity analysis, and fragment matching. Section~\ref{sec:results} presents the empirical results from our similarity analysis and provides an assessment of the efficacy of the proposed matching process. Finally, Section~\ref{sec:conclusion} and \ref{sec:future_directions} offer concluding remarks, discuss the inherent limitations of the current research, and suggest potential directions for future investigation in the field of generative multimedia carving.

\section{RELATED WORK}\label{sec:rw}
Early file carving techniques in digital forensics focused primarily on signature-based recovery, utilizing identifiable patterns such as headers and footers to detect and reassemble files from raw disks \cite{b2}. While effective in controlled or minimally fragmented environments, this approach becomes ineffective when files lack distinct signatures or when fragments are misaligned or interleaved \cite{b4}. The technical challenge is further exacerbated in the case of multimedia files, which exhibit high entropy and complex or compound encoding schemes, making them significantly harder to classify based on signatures alone. Consequently, researchers have turned to \textit{content-based} and \textit{machine learning-based} approaches to classify and recover file fragments without relying on metadata.

To address these challenges, Sportiello and Zanero proposed a context-based classification framework that integrates the classification results of neighboring fragments \cite{b7}. Their system utilizes support vector machines (SVMs) trained on a suite of statistical features, including entropy, byte frequency distribution (BFD), rate of change (RoC), and Lempel-Ziv complexity. By operating on the heuristic that adjacent fragments are likely to originate from the same file, their methodology significantly enhanced classification reliability—particularly for multimedia files, which often embed content from multiple formats.

Additionally, hierarchical classification has emerged as a promising machine learning approach for multimedia file carving. Qiu et al. proposed a hierarchical classification method that integrates statistical and supervised learning methods for classification and reassembly \cite{b3}. Their pipeline utilizes byte frequency distribution (BFD) and rate of change (RoC) features to train a support vector machine (SVM) classifier for file type detection. For the reassembly phase, they introduced the Parallel Unique Path (PUP) algorithm, which reconstructs fragmented files by matching sequences based on structural continuity. This methodology demonstrated superior performance compared to traditional carving tools, such as PhotoRec, in terms of recovery accuracy.

Similarly, Bhatt et al. developed a hierarchical SVM-based system that organized file types into taxonomies, with general classifiers at higher levels feeding into fine-grained classifiers at lower levels\cite{b4}. Evaluated on 14 file types from the Digital Corpora dataset, their model achieved a macro-averaged F1-score of 66\%. While their model performed well on simpler file types like CSV and TXT, accuracy declined for multimedia files, which contain compression formatting techniques such as Discrete Cosine Transform (DCT).

In terms of deep learning, Mittal et al. introduced FiFTy, a CNN-based classifier designed to classify file fragments using raw byte input\cite{b5}. By learning directly from byte sequences, FiFTy eliminates the need for hand-crafted features and achieved 77\% classification accuracy across 75 file types, demonstrating strong classification performance for large-scale environments. However, like other machine learning systems, FiFTy remains discriminative, able to identify fragment types, but lacking the capability to predict or reconstruct missing data.

Despite the significant progress achieved by these file carving methods, they continue to exhibit several common limitations. Most notably, their machine learning models are discriminative, focused on identifying fragment types but do not generate or predict missing fragments. This limits their effectiveness in situations where essential fragments of a file are lost or where reconstruction is dependent on partial input. Furthermore, these methods are limited under severe fragmentation conditions, where there is little structural information to guide recovery. These limitations inspired us to adopt generative modeling-based carving frameworks that can learn underlying binary patterns and generate plausible continuations of missing data.

\section{PRELIMINARIES}\label{sec:preliminaries}
We first introduce the bGPT \cite{b6}, a byte-level generative transformer that enables format-agnostic data generation by operating directly on raw binary sequences. Following this, we define the statistical and perceptual metrics used to quantify the fidelity of the predicted fragments. By integrating both byte-level distributional analysis and pixel-level structural measures, we provide a comprehensive assessment of the model’s ability to reconstruct forensically relevant data.

\subsection{bGPT}
\begin{figure}[htbp]
\centering
\includegraphics[width=\linewidth]{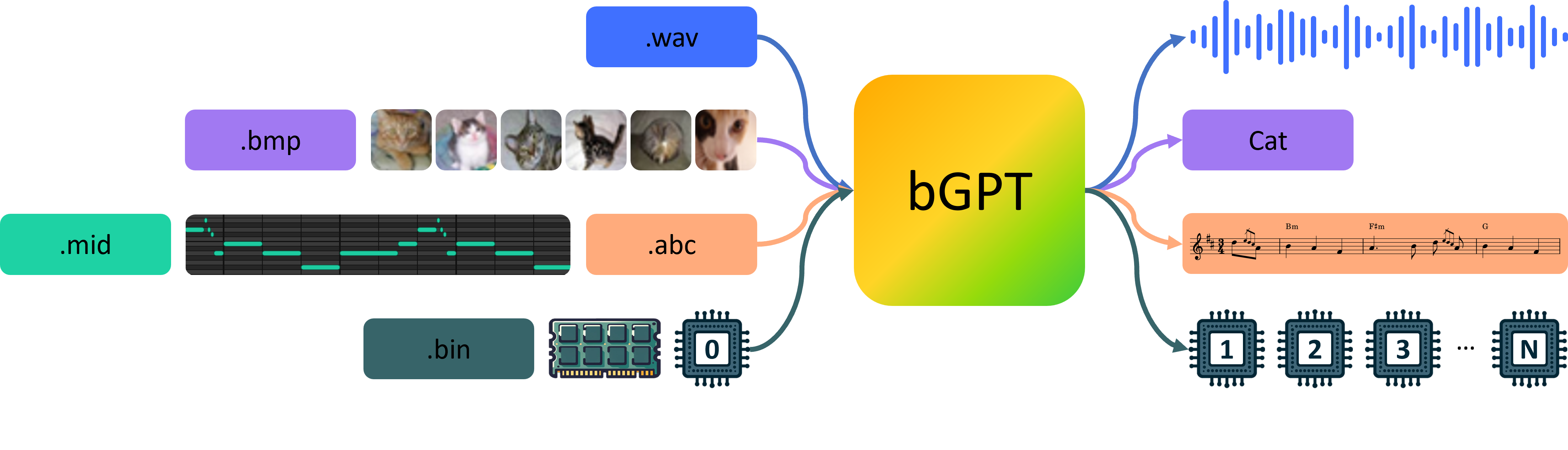}
\caption{The bGPT framework simulates digital systems through native binary data and integrates diverse data types into a single model, treating everything as a byte sequence.~\cite{b6}.}
\label{fig:bGPT}
\end{figure}

We utilize bGPT, an open-source, byte-level generative transformer developed for autoregressive byte generation. Originally, bGPT was tested on a wide range of generative tasks, including text generation, image generation, music conversion, and CPU behavior simulation, shown in Fig.~\ref{fig:bGPT}. The key strength of bGPT lies in its unified modeling capability, which enables diverse data types to be represented within a single framework by operating directly on raw bytes. This design enables generalization across domains and eliminates the need for format-specific metadata. In the context of our work, such flexibility is particularly valuable, as multimedia carving frequently encounters data with missing format information or fragmented structure.

In our implementation, we adopt the pre-trained \textit{image-weight} provided with the official bGPT model. Since this model was trained on $32\times32$ BMP images from the \textit{ImageNet} dataset\cite{b8}, the scope of our current study is limited to this configuration. However, the proposed approach is general and can be readily extended to images of arbitrary resolution by training the bGPT model on the corresponding image size. In our experiments, we slice partial fragments (2/5, 3/5, or 4/5) of a BMP file and input these byte sequences into bGPT to generate predicted continuations. The generated outputs are then compared against the real continuation of the original files using statistical similarity metrics. Through this setup, we evaluate bGPT’s capacity to reconstruct byte-level content for fragment matching in multimedia forensic applications.

\subsection{Similarity Metrics}

To evaluate byte-level and pixel-level similarity between a predicted BMP image fragment and its real continuation, we apply four statistical metrics: Cosine Similarity, Chi-Square Distance, Jensen-Shannon Divergence, and Structural Similarity Index Measure.

\textbf{1) Cosine Similarity [0--1]:} Cosine similarity measures the angle between two non-zero vectors, capturing how closely their directions align\cite{b9}. In fragment comparison, it evaluates whether the predicted and actual byte frequency histograms share a similar distribution pattern. 
\begin{equation}
\cos(\theta) = \frac{A \cdot B}{\|A\| \|B\|}
\end{equation}
Here, $A$ and $B$ are byte frequency vectors for the predicted and real fragment, $A \cdot B$ is the dot product, and $\|A\|$, $\|B\|$ are their vector magnitudes. A value close to 1 indicates strong compositional alignment between byte distributions.

\textbf{2) Chi-Square Distance [0--$\infty$):} Chi-square distance measures the difference between two distributions by comparing their frequency counts\cite{b9}. In fragment comparison, it quantifies how byte frequency distributions between the predicted and real fragments differ, being sensitive to frequency variations and scale differences.
\begin{equation}
\chi^2 = \sum \frac{(O_i - E_i)^2}{E_i}
\end{equation}
Here, $O_i$ is the observed byte frequency from the predicted fragment, and $E_i$ is the expected byte frequency from the real fragment. Smaller $\chi^2$ values indicate smaller differences in the frequencies of individual byte values between the distributions.

\textbf{3) Jensen-Shannon Divergence (JSD) [0--1]:} JSD is a symmetric and smoothed variant of Kullback-Leibler divergence that measures the similarity between two probability distributions\cite{b9}. In fragment comparison, it evaluates how predicted and real byte fragments diverge as a probability distribution, remaining stable in presence of zero bins.
\begin{equation}
\text{JSD}(P \parallel Q) =
\frac{1}{2} D_{\text{KL}}(P \parallel M) +
\frac{1}{2} D_{\text{KL}}(Q \parallel M)
\end{equation}
Here, $M$ is the average distribution between $P$ and $Q$, and $D_{\text{KL}}$ is the Kullback-Leibler divergence. JSD values close to 0 indicate less divergence and stronger similarity between byte distributions.

\textbf{4) Structural Similarity Index Measure (SSIM) [0--1]:} SSIM is a perceptual-based metric developed to measure the structural similarity between two images\cite{b10}. In fragment comparison, it evaluates pixel-level consistency between the predicted and real image fragments.
\begin{equation}
\text{SSIM}(x, y) =
\frac{(2\mu_x \mu_y + C_1)(2\sigma_{xy} + C_2)}
{(\mu_x^2 + \mu_y^2 + C_1)(\sigma_x^2 + \sigma_y^2 + C_2)}
\end{equation}
Here, $\mu_x$ and $\mu_y$ are the mean pixel intensities, $\sigma_x^2$ and $\sigma_y^2$ are the variances, and $\sigma_{xy}$ is the covariance between fragments. SSIM values close to 1 indicate stronger pixel-wise structural similarity between predicted and real fragments.

\section{METHODOLOGY}
\label{sec:methodology}

\subsection{Phase 1: Dataset Preparation}

In our experiments, we utilize a subset of the \textit{ImageNet} validation dataset. From this dataset, we randomly select 2250 images and convert each image into $32 \times 32$ BMP format. To simulate partial fragmentation, the selected images are divided into three separate input sets:

\begin{itemize}
    \item \textbf{$I_1$:} 750 BMP images sliced to retain 2/5 of their full byte length.
    \item \textbf{$I_2$:} 750 BMP images sliced to retain 3/5 of their full byte length.
    \item \textbf{$I_3$:} 750 BMP images sliced to retain 4/5 of their full byte length.
\end{itemize}

Further, each input set consists of three subsets:

\begin{itemize}
    \item \textbf{Full Images:} Contains the original $32 \times 32$ BMP images ($\sim$ 3.2 kB) in their entirety.
    \item \textbf{Input Fragments:} Contains the partially sliced images ($I_1$, $I_2$, $I_3$) used as model input.
    \item \textbf{Real Fragments:} Contains the true remaining byte fragments cut from the original images, representing the ground truth continuations.
\end{itemize}

\subsection{Phase 2: Generative Prediction with bGPT}

We utilize bGPT to generate predicted byte continuations. Each input fragment --- $I_1$, $I_2$, and $I_3$ --- is independently fed into the model. The model processes each partial input and generates a predicted byte continuation based on the learned structure from the pre-trained \textit{image-weights}. To organize the outputs, we store the generated predictions into separate sets according to the fragment size used for input:

\begin{itemize}
    \item \textbf{$P_1$ ($\sim$ 1.8 kB):} Contains the predicted fragment continuations when the input was $I_1$.
    \item \textbf{$P_2$ ($\sim$ 1.2 kB):} Contains the predicted fragment continuations when the input was $I_2$.
    \item \textbf{$P_3$ ($\sim$ 0.6 kB):} Contains the predicted fragment continuations when the input was $I_3$.
\end{itemize}

\subsection{Phase 3: Similarity Analysis}

To evaluate the quality of the predicted fragments, we conduct a similarity analysis between each predicted fragment and its corresponding real continuation across different input sizes. For byte-level comparison, we calculate three statistical similarity metrics: chi-square distance, cosine similarity, and JSD. For pixel-level similarity, we apply SSIM to assess pixel-wise structural similarity between the predicted and real BMP image fragments.

Based on the similarity analysis of each fragment, we compute statistical summaries for each prediction set (i.e., $P_1$, $P_2$, and $P_3$), including the mean, median, minimum, maximum, standard deviation, and 95\% confidence interval error margin ($\pm$) of the similarity scores. The results from the byte-level similarity analysis serve as the basis for designing a matching function. This function is used to identify and match predicted fragments to their corresponding real continuations during the fragment matching phase.

\begin{figure}[htbp]
\centering
\includegraphics[width=0.8\linewidth]{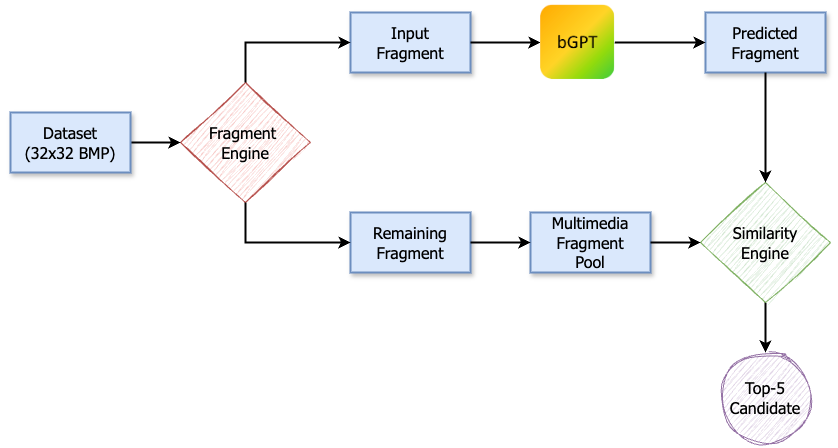}
\caption{Flow diagram of a pseudo-simulation pipeline where fragmented BMP inputs are processed by bGPT to predict the next fragment, which is compared to a mixed multimedia fragment pool using similarity metrics to identify the top five matches.}
\label{fig:fragmatch}
\end{figure}

\subsection{Phase 4: Fragment Matching}

For fragment matching, we use a set of 30 randomly selected predicted BMP fragments, 10 fragments from each input ($I_1$, $I_2$, $I_3$), to evaluate the matching process. Each predicted fragment, as shown in Fig.~\ref{fig:fragmatch}, is compared against a multimedia fragment pool consisting of 100 real fragments of the same size, including its actual BMP continuation and fragmented WAV, JPEG, PNG, and MP4 files.

To rank matches, each candidate fragment in the pool is given a score based on a weighted function:

\begin{equation}
S = (\alpha \times \text{Chi}) + (\beta \times \text{JSD}) - (\gamma \times \text{Cos})
\end{equation}

where $\alpha$, $\beta$, and $\gamma$ are tunable weights corresponding to the chi-square distance, JSD, and cosine similarity results. Lower scores indicate greater similarity to the predicted fragment. For each prediction, the top five candidate fragments with the lowest scores are selected. The results demonstrate how well the generative model's predicted fragment aligns with its real continuation against other fragmented multimedia files.

\section{RESULTS}\label{sec:results}

\subsection{Similarity Analysis Across Fragment Sizes}

The results in Table~\ref{tab:stats} statistically quantify the similarity between bGPT’s predicted fragments (i.e., $P_1$, $P_2$, and $P_3$) and their real continuations across different input sizes. It can be seen that smaller predictions generated from larger input fragments consistently achieve higher similarity scores across most of the evaluated metrics.

\begin{table*}[htbp]
\centering
\caption{Results of similarity analysis between predicted fragments and their real continuations across fragment sizes.}
\label{tab:stats}
\renewcommand{\arraystretch}{1.9}
\setlength{\tabcolsep}{8pt}
\resizebox{\textwidth}{!}{
\begin{tabular}{|c|c|c|c|c|c|c|c|c|c|}
\hline
\multirow{2}{*}{\textbf{Metric}} & \multicolumn{3}{c|}{\textbf{Mean}} & \multicolumn{3}{c|}{\textbf{Std Dev}} & \multicolumn{3}{c|}{\textbf{95\% CI (±)}} \\
\cline{2-10}
& $P_1$ ($\sim$ 1.8 kB) & $P_2$ ($\sim$ 1.2 kB) & $P_3$ ($\sim$ 0.6 kB)
& $P_1$ ($\sim$ 1.8 kB) & $P_2$ ($\sim$ 1.2 kB) & $P_3$ ($\sim$ 0.6 kB)
& $P_1$ ($\sim$ 1.8 kB) & $P_2$ ($\sim$ 1.2 kB) & $P_3$ ($\sim$ 0.6 kB) \\
\hline
Chi-square & 409.58 & 335.61 & 174.97 & 161.23 & 139.24 & 61.96 & 23.12 & 19.96 & 8.88 \\
\hline
Cosine     & 0.6690 & 0.6749 & 0.7641 & 0.1735 & 0.1605 & 0.1198 & 0.0249 & 0.0230 & 0.0172 \\
\hline
JSD        & 0.1864 & 0.2338 & 0.2458 & 0.0798 & 0.1050 & 0.0938 & 0.0114 & 0.0151 & 0.0134 \\
\hline
SSIM       & 0.0856 & 0.1016 & 0.3262 & 0.0738 & 0.0966 & 0.1628 & 0.0106 & 0.0138 & 0.0233 \\
\hline
\end{tabular}}
\end{table*}

\begin{figure*}[t]
\centering
\includegraphics[width=\textwidth,height=0.2\textheight]{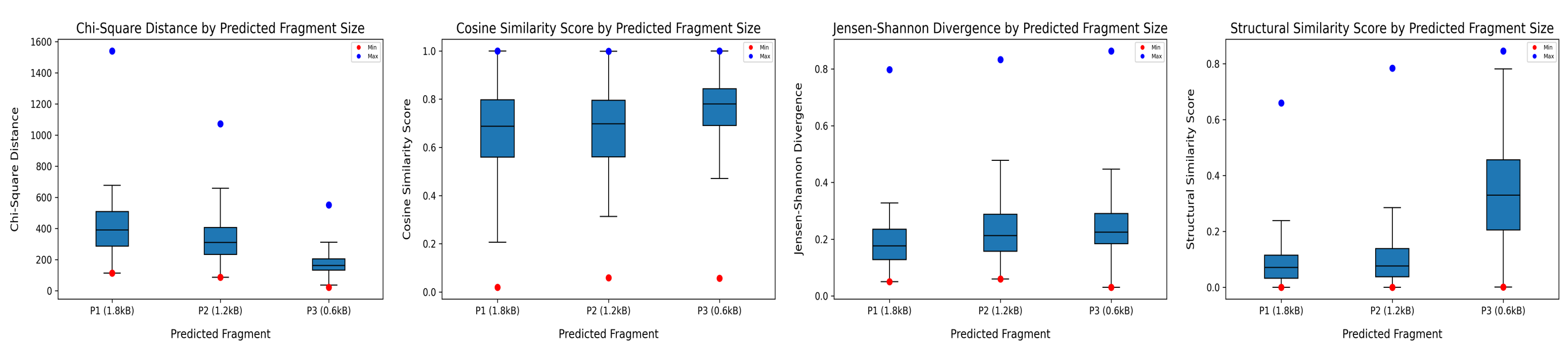}
\caption{Box plots showing similarity scores between predicted and real fragments across three predicted fragment sizes: $P_1$ (1.8 kB), $P_2$ (1.2 kB), and $P_3$ (0.6 kB), using four evaluation metrics: cosine similarity, SSIM, chi-square distance, and Jensen--Shannon divergence. Each plot captures the distribution and variation in similarity across different prediction groups.}
\label{fig:boxplots}
\end{figure*}

Fig.~\ref{fig:boxplots} depicts the similarity trends from bGPT predictions across varying input sizes. While there exists variability in prediction quality, with some predicted fragments achieving near-perfect similarity, the overall trend shows improvement in similarity and consistency given larger inputs. The extreme values across metrics show that bGPT can occasionally produce hallucinated, low-quality outputs. Despite such occasional invalid predictions, the aggregate findings show that bGPT is capable of producing predictive continuations that align closely with real fragments at the byte level.

\subsection{Byte-level analysis}

The mean cosine similarity improved from 0.6690 in $P_1$ to 0.7641 in $P_3$ (Table~\ref{tab:stats}), indicating stronger proportional alignment of byte-level structures when larger input fragments were provided. Because cosine similarity captures the shape of the byte distribution, bGPT’s performance reveals that it effectively preserves the global structure of the original file when predicting unseen data. The overall high scores demonstrate that bGPT maintains consistency in distributional patterns, allowing predicted fragments to integrate naturally with existing data without introducing significant structural anomalies. Moreover, the relatively low standard deviation and narrow 95\% confidence interval indicate that bGPT’s predictions are structurally reliable and stable across different fragment inputs.

Similarly, the mean chi-square distance improved from 409.58 in $P_1$ to 174.97 in $P_3$ (Table~\ref{tab:stats}), indicating that byte frequency discrepancies between predicted and real distributions were significantly reduced with more input data. Since chi-square distance is sensitive to localized variations in byte frequency, bGPT demonstrates improved byte-level precision in reproducing local frequency patterns. Additionally, the lower standard deviation and narrower 95\% confidence interval for $P_3$ indicate increased stability and fewer inconsistencies at the byte level.

Conversely, the mean Jensen--Shannon divergence (JSD) increased from 0.1864 in $P_1$ to 0.2458 in $P_3$ (Table~\ref{tab:stats}). One possible explanation is that larger input fragments provide richer context, allowing the model to generate continuations with greater structural or statistical diversity. Such diversity can increase the distributional spread of the predicted fragments and, consequently, produce higher JSD values. In this sense, the increase in JSD may reflect greater variability in the generated content.

\subsection{Pixel-level analysis}

\begin{figure}[t]
\centering
\includegraphics[scale=.5]{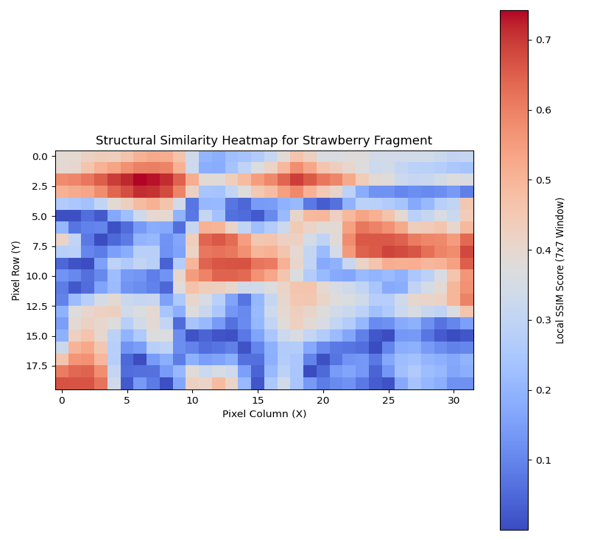}
\caption{SSIM heatmap between predicted and real image fragments from Fig.~\ref{fig:strawberry}. Local SSIM values were computed using a $7\times7$ sliding window, where each pixel represents the similarity of its local neighborhood.}
\label{fig:heatmap}
\end{figure}

The mean structural similarity index measure (SSIM) improved, increasing from 0.0856 in $P_1$ to 0.3262 in $P_3$ (Table~\ref{tab:stats}). Although the overall SSIM values remain relatively low compared to traditional image similarity standards, the trend reveals meaningful gains in perceptual structure as input size increases. Importantly, the local SSIM heatmap in Fig.~\ref{fig:heatmap} shows that even when global SSIM values are low, bGPT’s predictions exhibit localized regions of strong pixel-level consistency, particularly around edges and texturally uniform regions. This behavior indicates that bGPT can reconstruct meaningful visual features even when broader pixel patterns vary. In multimedia forensic applications, preserving localized structure is particularly important, as analysts often rely more on recognizable visual features than on exact pixel-by-pixel reconstruction.

\begin{figure}[b]
\centering
\includegraphics[scale=.5]{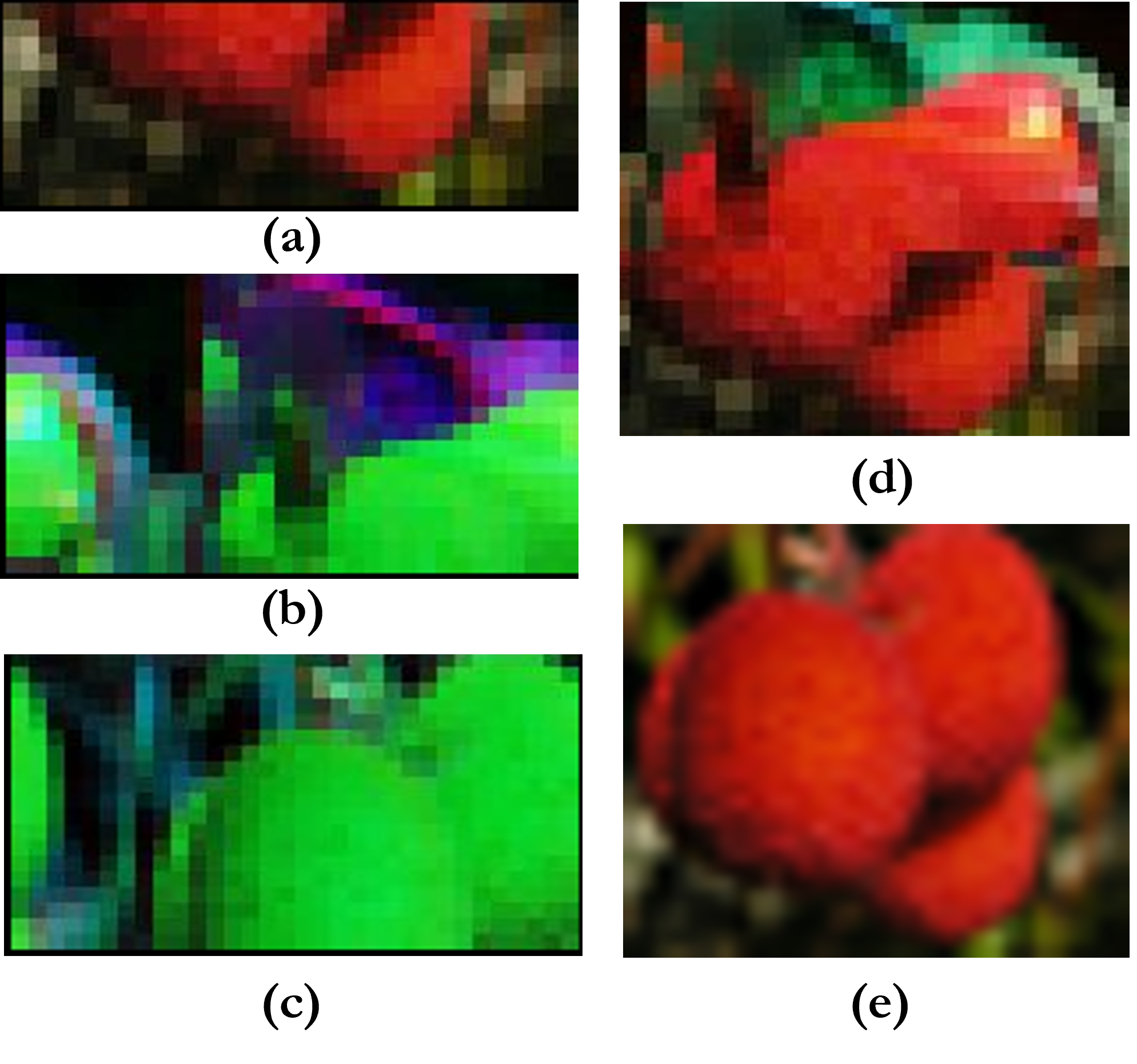}
\caption{Example of fragment reconstruction using bGPT. The input fragment (a) is provided to the model to generate the predicted continuation (b), which approximates the true continuation (c). Combining (a) and (b) produces the reconstructed image (d), compared against the original full image (e).}
\label{fig:strawberry}
\end{figure}

The perceptual analysis of the reconstructed strawberry fragment in Fig.~\ref{fig:strawberry} demonstrates that the predicted continuation preserves color regions, object contours, and overall image composition, despite minor inconsistencies in fine-texture details. Although small artifacts and misalignments are present, the overall perceptual coherence indicates that bGPT’s predictions retain essential visual characteristics relevant for forensic interpretation. These findings suggest that bGPT’s generative outputs are not only statistically consistent but also perceptually meaningful under limited input conditions.

\subsection{Matching Based on Statistical Analysis}

From the average scores of the similarity metrics, we assigned matching weights of $\alpha = 0.01$, $\beta = 10$, and $\gamma = 10$, resulting in the following scoring function:

\begin{equation}
S = (0.01 \times \text{Chi}) + (10 \times \text{JSD}) - (10 \times \text{Cos})
\end{equation}

Using this matching function, Table~\ref{tab:match} summarizes the fragment matching results. Out of 30 predicted fragments, 20 correctly ranked the real continuation as the top match. Additionally, 4 predictions placed the correct match within the top five candidates, with 3 of those ranking the real fragment second. However, 6 predictions failed to identify the correct continuation within the top five. 

Overall, these results indicate that bGPT, when combined with an effective similarity-based matching function, has strong potential to support digital forensic analysts in the matching and reassembly of multimedia fragments from raw disk data.

\begin{table}[htbp]
\centering
\caption{Fragment matching results.}
\label{tab:match}
\begin{tabular}{l c}
\hline
\textbf{Ranking Outcome} & \textbf{Count} \\
\hline
Correct match ranked 1st & 20 \\
Correct match within top 5 (not 1st) & 4 \\
Correct match not within top 5 & 6 \\
\hline
\textbf{Total predictions} & 30 \\
\hline
\end{tabular}
\end{table}

\section{CONCLUSION}\label{sec:conclusion}

This research explored the application of a byte-level generative model, bGPT, for multimedia fragment prediction. By providing partial BMP fragments as input, bGPT generated plausible byte-level continuations, which were evaluated using multiple similarity metrics. The results showed that larger input fragments consistently improved prediction quality, with cosine similarity, structural similarity index measure (SSIM), and chi-square distance indicating stronger alignment between predicted and real continuations.

Despite these promising results, several limitations remain. First, experimentation was restricted to BMP fragments due to the use of pre-trained image weights trained on $32\times32$ ImageNet data. Second, fragment generation was limited to fixed input sizes (2/5, 3/5, and 4/5 of the full image), whereas real-world forensic scenarios involve variable-length fragments. Additionally, the pre-trained bGPT architecture imposes an output limit of approximately 8 kB, restricting the generation of longer fragments.

\section{FUTURE DIRECTIONS}\label{sec:future_directions}

Future work will focus on extending this approach to additional multimedia formats such as JPEG, PNG, WAV, and MP4, enabling broader applicability in forensic environments. Incorporating format-aware features such as headers, footers, and file signatures into the generative process may improve reconstruction fidelity when metadata is absent. Furthermore, exploring variable-length fragment prediction, multi-fragment reconstruction, and methods to guide model attention toward fragment boundaries may enhance performance in more realistic fragmentation scenarios. These directions aim to further develop generative models as practical tools for multimedia forensic carving and reconstruction.

\bibliography{report} 
\bibliographystyle{spiebib} 

\end{document}